# An Interactive Multi-Agent System for Evaluation of New Product Concepts

Bin Xuan
Department of Data Science
Seoul National University of Science and Technology
Seoul, Republic of Korea
xuanbin159@seoultech.ac.kr

Ruo Ai
TCL China Star Optoelectronics Technology Co., Ltd.
Hubei, China
jxytair@163.com

Hakyeon Lee*
Department of Industrial Engineering
Seoul National University of Science and Technology
Seoul, Republic of Korea
hylee@seoultech.ac.kr

**Abstract**

Product concept evaluation is a critical stage that determines strategic resource allocation and project success in enterprises. However, traditional expert-led approaches face limitations such as subjective bias and high time and cost requirements. To support this process, this study proposes an automated approach utilizing a large language model (LLM)-based multi-agent system (MAS). Through a systematic analysis of previous research on product development and team collaboration, this study established two primary evaluation dimensions, namely technical feasibility and market feasibility. The proposed system consists of a team of eight virtual agents representing specialized domains such as R&D and marketing. These agents use retrieval-augmented generation (RAG) and real-time search tools to gather objective evidence and validate concepts through structured deliberations based on the established criteria. The agents were further fine-tuned using professional product review data to enhance their judgment accuracy. A case study involving professional display monitor concepts demonstrated that the system's evaluation rankings were consistent with those of senior industry experts. These results confirm the usability of the proposed multi-agent-based evaluation approach for supporting product development decisions.

---

* Corresponding authors.

# 1. Introduction

Product concept evaluation represents an essential decision-making stage in modern enterprise product development. This stage directly influences strategic resource allocation for innovation and ultimately determines project success and return on investment (Haefner et al., 2021). This significance becomes particularly pronounced in an increasingly competitive global market characterized by shortened product lifecycles, where effective concept evaluation not only helps enterprises mitigate risks and avoid resource waste, but also enables them to seize market opportunities and build long-term competitive advantages (Cooper & Kleinschmidt, 1993). However, contemporary evaluation processes face complex challenges that demand increasingly sophisticated approaches. These processes must carefully consider the interactions and potential conflicts among multiple dimensions including market analysis, technical feasibility, user needs, and supply chain resilience (Gama & Magistretti, 2025). Rapidly changing market environments further demand higher efficiency and responsiveness in evaluation processes to match the pace of agile development (de Borba et al., 2019).

Existing product concept evaluation methods have undergone a developmental trajectory from traditional expert judgment to intelligent evaluation. Traditional evaluation approaches rely on expert judgment as their core and employ qualitative or semi-quantitative methods such as the delphi method, analytic hierarchy process (AHP), and focus groups (Bustinza et al., 2024; Cooper, 2019). Although these methods are widely applied in practice, their inherent limitations have become increasingly pronounced in today's environment. Evaluation results are susceptible to individual cognitive biases that undermine the objectivity and consistency of conclusions (Yehudai et al., 2025). Moreover, the knowledge boundaries of any expert group are inherently limited, which poses particular challenges when evaluating interdisciplinary innovative concepts (Eriksson et al., 2025). Organizing and coordinating expert reviews also demands considerable time. This creates evaluation cycles that conflict with the pursuit of rapid iteration in modern product development (Zhao & Liu, 2024).

With the advancement of artificial intelligence technology, evaluation methods based on large language models (LLMs) have begun to emerge. LLMs have acquired powerful language understanding, knowledge reasoning, and content generation capabilities through pre-training on massive text datasets (Brown et al., 2020; Bubeck et al., 2023). Beyond these core capabilities, LLM-based agents extend the role of LLMs from passive question-answering tools to autonomous systems that can perceive environments, formulate plans, and invoke external tools (Yao et al., 2023). When multiple such agents are organized into a multi-agent system (MAS), each assigned a distinct professional role, they can simulate cross-functional expert committees that conduct multi-perspective evaluation through structured collaboration and debate at low cost and large scale (Park et al., 2023).

Despite this promising potential, current LLM agent research remains predominantly concentrated on well-defined domains such as software development, code generation, and structured question answering. These domains share common characteristics such as clear task boundaries and easily measurable success criteria. In contrast, product concept evaluation presents fundamentally different challenges that require careful judgment across multiple interdisciplinary dimensions and the synthesis of diverse stakeholder perspectives. To date, no studies have attempted to apply LLM agent systems for this complex evaluation task, leaving a significant gap between the demonstrated capabilities of these intelligent systems and their practical application in strategic innovation decision-making. Nevertheless, recent advances in multi-agent architectures that enable collaborative reasoning and structured deliberation now facilitate a new approach to product concept evaluation.

To address this gap, this study proposes an LLM-based MAS for new product concept evaluation in which LLM agents simulate cross-functional teams to execute the evaluation. The proposed approach evaluates product concepts by assigning fixed professional roles to agents and guiding structured discussions. This study establishes three specific research objectives. The first objective is to develop a systematic evaluation framework through thorough literature review of product development and team collaboration research, defining evaluation dimensions with their constituent criteria and specifying cross-functional agent team composition. The second objective is to implement the technical architecture by designing a MAS that integrates retrieval-augmented generation (RAG) and tool to enable dynamic knowledge acquisition and coordinated agent collaboration. The third objective is to validate the proposed approach through product concept case studies that demonstrate the system's practical applicability and evaluation quality compared to traditional evaluation methods.

The remainder of this paper proceeds as follows. Section 2 examines previous research on product concept evaluation and LLM-based agent technology. Section 3 presents the proposed system design with its evaluation criteria, agent roles, and workflow architecture. Section 4 demonstrates the system through a case study of professional display monitor concepts, details the agent fine-tuning process, and assesses evaluation quality by comparing multi-agent results against human expert ratings. Section 5 discusses practical challenges encountered during implementation. Section 6 concludes the study and suggests directions for future research.

## 2. Related works

*2.1. Previous approaches to evaluation of new product concepts*

Product concept evaluation serves as a critical decision-making stage in new product development (NPD), with the core objective of evaluating the feasibility, market potential, and commercial value of concepts before formal product development begins (Cooper & Kleinschmidt, 1993). Evaluation content primarily covers dimensions including technical feasibility, market potential, commercial value, and risk assessment. With technological advancement, product concept evaluation approaches have undergone an evolutionary process from qualitative to quantitative, from subjective to objective, and from unidimensional to multidimensional. Research by Hart et al. (2003) demonstrates that effective product concept evaluation can significantly improve new product success rates and reduce product development failure risks. This evolution of evaluation approaches reflects enterprises' continuous pursuit of improved decision-making rigor, reduced innovation risks, and accelerated product time-to-market.

Traditional expert evaluation approaches are based on human experts' knowledge and experience, achieving systematic evaluation of product concepts through structured processes. In practice, simple scoring methods where experts rate concepts on scales have been widely used, but to overcome the limitations of such straightforward scoring, multi-criteria decision making (MCDM) methods have emerged as the mainstream approach in product concept evaluation. Primary MCDM methods include the AHP, technique for order of preference by similarity to ideal solution (TOPSIS), and weighted scoring methods (Mardani et al., 2015). The effectiveness of such systematic expert evaluation has been empirically validated, with Cooper & Kleinschmidt (1993) demonstrating high accuracy in predicting product success rates, thereby providing important support for early decision-making. Building upon these foundational methods, Calantone et al. (1999) further developed multidimensional evaluation models encompassing product characteristics, market characteristics, and technical characteristics.

Despite these methodological advancements, traditional expert evaluation approaches have significant limitations. Subjectivity bias is the most prominent problem, as experts' personal experience and cognitive biases affect the objectivity of evaluation results (Schmidt et al., 2001). High time costs severely constrain application efficiency, as organizing expert reviews typically requires lengthy cycles that conflict with agile development demands (Verworn et al., 2008). Scalability limitations make traditional approaches difficult to apply to large quantities of product concepts. Lagged knowledge updates result in experts potentially making evaluations based on outdated technical cognition, particularly in emerging technology fields where expert experience value significantly decreases (Magnusson, 2009). Additionally, poor result consistency and difficulties in cross-domain integration further limit the widespread application of traditional approaches.

To overcome the limitations of traditional approaches, machine learning techniques have been integrated into product concept evaluation processes. Topic modeling methods such as LDA (Blei et al., 2003) enabled automated extraction of product attributes from large volumes of documents and user reviews (Joung & Kim, 2021), while word embedding models such as Word2Vec (Mikolov et al., 2013) facilitated semantic similarity-based concept classification. More recently, BERT (Devlin et al., 2019) advanced contextual understanding of product reviews and feature-based scoring tasks. These techniques collectively improved evaluation efficiency and enabled large-scale processing of product concepts.

However, these approaches exhibit fundamental limitations that constrain their applicability to product concept evaluation. They can extract features, calculate similarities, and classify concepts. However, they cannot independently perform integrated evaluations that combine technical feasibility, market potential, and strategic fit into actionable recommendations (Witkowski & Wodecki, 2025). In terms of output, these methods are limited to classification or scoring results and cannot generate evaluation reports that include reasoning and improvement suggestions (Yadav & Vishwakarma, 2020). Their semantic understanding also relies primarily on pattern matching rather than grasping the commercial logic and value propositions behind product concepts (Cambria & White, 2014). Furthermore, model performance degrades substantially in cross-domain applications, requiring data recollection and retraining for each new domain (Pan & Yang, 2010; Zhang et al., 2022). These deficiencies motivated the exploration of LLM-based approaches that combine broad knowledge with reasoning and generation capabilities.

In recent years, the emergence of LLMs has provided new approaches to address these limitations. LLMs possess deep semantic understanding capabilities, enabling understanding of the innovative connotations of product concepts and grasping the value propositions and commercial logic behind concepts, which addresses the pattern matching limitations of prior machine learning methods (Brown et al., 2020). This deep understanding capability enables LLMs to handle complex commercial contexts and implicit meanings. The Chain-of-Thought technology proposed by Wei et al. (2023) further enables LLMs to provide transparent reasoning processes, enhancing the interpretability of evaluation results. Generative capabilities directly address the output limitations of prior methods, enabling generation of detailed evaluation reports, provision of specific improvement suggestions, and explanation of evaluation logic (Ouyang et al., 2022). The few-shot learning capability addresses the data dependency problem, requiring only a few examples to adapt to new evaluation tasks through broad knowledge obtained from pre-training (Brown et al., 2020). Multimodal processing capabilities enable simultaneous analysis of text, images, tables, and other forms of product concept information (Radford et al., 2021), and dynamic adaptability allows LLMs to adjust evaluation strategies based on real-time feedback, supporting progressive refinement of product concepts (Zhao & Liu, 2024). These capabilities collectively suggest that LLMs can serve as a foundation for more autonomous product concept evaluation systems.

*2.2. LLM-based multi-agents*

A significant paradigm shift is occurring in the field of artificial intelligence, from passive responsive models to proactive, goal-oriented intelligent agents. These agents driven by LLMs can perceive environments, reason, and execute actions, representing a potential path toward artificial general intelligence (Luo et al., 2025). In academic literature, research on LLM agents can be examined at two levels. The first level analyzes single agents as independent operating units, while the second level studies collective systems when multiple agents collaborate to solve complex problems. This agentification transformation represents not only an upgrade in technical architecture, but also a shift toward artificial intelligence playing a more active and strategic role in innovation processes, increasingly supporting and augmenting human decision-making (Mariani et al., 2023; Modgil et al., 2025).

The essence of a single LLM agent is not the LLM itself, but rather a structured system with the LLM as the core controller, enhanced through key components such as planning, memory, and tool use. Planning is the core capability of agents, involving decomposing large complex tasks into smaller, more manageable sub-goals, and iteratively improving behavior through self-reflection (Cheng et al., 2024; Wang et al., 2024). Memory components aim to overcome the fundamental limitation of LLMs lacking persistent memory between interactions, while tool use endows agents with the ability to access external information by calling external APIs. The effective operation of these advanced components heavily relies on RAG as a core technology, which grounds agents in verifiable reality by retrieving information from external knowledge bases (Neha et al., 2025). In the application context of product evaluation, single LLM agents have demonstrated unique cognitive advantages. Research by Boiko et al. (2023) significantly enhanced agents' reasoning capabilities in scientific research by integrating domain-specific tools for single agents, providing new possibilities for technical feasibility analysis of product concepts. The latest research by Collier et al. (2025) shows that LLMs perform excellently in divergent thinking tasks for product risk evaluation, identifying potential risks and opportunities that human experts might overlook.

However, existing research also reveals systematic limitations of the single-agent paradigm in complex evaluation tasks. As task complexity increases, agents often encounter cognitive overload and confusion during tool selection, leading to suboptimal or even erroneous results (Raza et al., 2025). Furthermore, single agents demonstrate remarkably limited self-correction and validation capabilities, finding it difficult to reliably verify the effectiveness of their own reasoning (Stechly et al., 2024). Most fundamentally, single agents are essentially isolated rational actors operating from a single, homogeneous knowledge base. Consequently, they are unable to truly replicate the diverse perspectives and collaborative debate necessary for solving complex problems, making them inherently less suitable than MASs for tasks requiring multiple perspectives. These limitations are particularly pronounced in complex decision-making scenarios, such as product concept evaluation, which require multi-stakeholder perspectives.

To break through the systematic bottlenecks of single agents, research has turned to MAS, solving complex problems by decomposing cognitive burden among multiple specialized agents (Brea, 2023). The core advantages of MAS lie in their task-solving potential, system flexibility, and ability to integrate specialized knowledge (Yang et al., 2024). The concept of MAS originates from traditional distributed artificial intelligence, but with the emergence of LLMs, inter-agent interactions have evolved from structured data exchange to more complex natural language communication. As system complexity increases, effectively managing inter-agent communication and shared context becomes a core challenge, spurring research on model context protocols that establish rules and mechanisms ensuring agents can effectively synchronize information and coordinate actions in distributed environments.

The research development trajectory of MAS clearly reflects the evolution of its application focus. Early pioneering work studied how MASs generate emergent social dynamics by simulating believable human behavior in interactive sandbox environments, typically featuring sophisticated memory and reflection architectures enabling planning based on long-term experience (Park et al., 2023; Wu et al., 2023). Subsequently, research focus shifted to more structured collaboration patterns, such as coordinating a group of agents playing different software engineering roles through encoding standard operating procedures to efficiently complete complex programming tasks (Hong et al., 2024). Recent research trends have further developed toward more general, flexible frameworks, such as AutoGen and AgentVerse systems that allow developers to create customized multi-agent workflows for various application scenarios (Chen et al., 2023; Wu et al., 2023).

Critical examination of these mainstream frameworks reveals a common limitation. Current MAS research is highly concentrated in programmatic and data-driven domains, while significant gaps remain in other professional fields requiring simulation and analysis of subjective human judgment, such as marketing, product design, and strategic management (Raza et al., 2025). Product concept evaluation, as a key component of the fuzzy front end of business innovation, precisely embodies the gaps existing in current agent technology paradigms (Singh et al., 2024). Product concept evaluation is a complex problem domain that integrates multiple challenges including social simulation, structured analysis, and creative iteration. Research by Huang et al. (2025) emphasizes that LLMs realize a new paradigm of human-machine collaboration, where through natural language interaction, evaluators can engage in deep dialogue with LLMs, clarifying evaluation criteria, exploring improvement directions, and verifying evaluation conclusions. An effective evaluation system must be able to instantiate and manage a group of agents representing different target user groups, generate rich qualitative feedback, and support both structured and unstructured interaction modes (Ingomar & Joo, 2012).

Single agent systems fundamentally lack the multi-perspective reasoning capabilities necessary for evaluation, while existing mainstream multi-agent frameworks are equally unsuitable. Systems aimed at simulating social behavior provide the behavioral authenticity needed for role simulation but lack goal-oriented frameworks to guide these roles in completing structured evaluation tasks. Conversely, structured workflows designed for tasks like software engineering are too rigid and programmatic for creative and subjective tasks like product concept evaluation. Therefore, the product concept evaluation field requires a newly designed MAS that simulates stakeholder ecosystems to provide iterative, multi-faceted feedback. To address this need, the following section presents the proposed approach for constructing such a system.

## 3. Proposed approach

*3.1. Overall approach*

This section elaborates in detail on the overall approach and core components for constructing a multi-LLM agent-based product concept evaluation system. The proposed approach adopts a systematic design that divides the entire process into two stages of evaluation model design and agent system construction. Figure 1 shows the overall framework of this research, and subsequent sections will provide detailed explanations of each module.

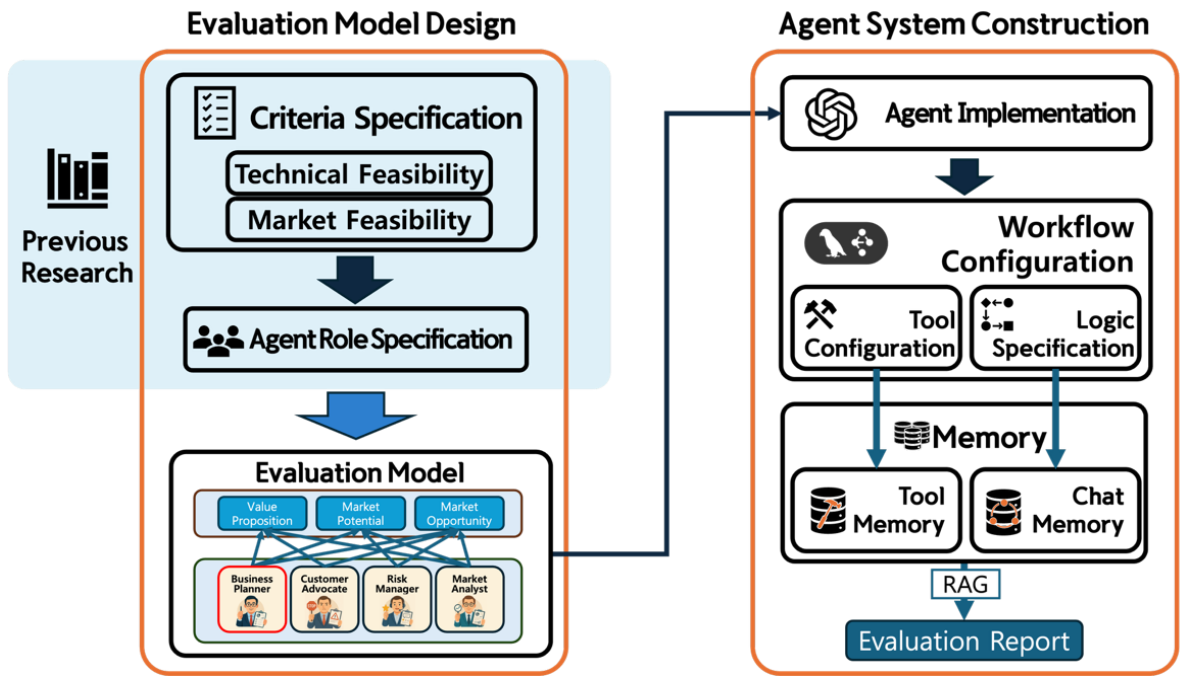

Fig. 1. Framework of the proposed approach

In the evaluation model design stage, the primary task is to construct a structured analysis framework. This process begins with in-depth analysis of previous research to establish detailed internal evaluation criteria for the two dimensions of technical feasibility and market feasibility. Simultaneously, a series of agent roles required for executing the evaluation are defined based on these specific criteria. The evaluation model is then formed through systematic linking of these clear evaluation benchmarks with corresponding agent roles. This integrated model provides clear theoretical guidance and structural foundation for subsequent automated evaluation.

In the subsequent agent system construction stage, the above conceptual model is implemented technically. LLM is selected as the primary technology for implementing each agent due to its powerful language understanding, reasoning, and role simulation capabilities. To enable these LLM agents to collaborate effectively, a sophisticated workflow is established that specifies the interaction sequence and logic among agents. Meanwhile, to overcome the limitations of static knowledge in LLMs, the system equips agents with the capability to invoke external tools, such as allowing access to real-time network data or executing code analysis. A key innovation of the system lies in its memory architecture, which possesses both chat memory for recording evaluation process context and tool memory for storing tool invocation results. To efficiently utilize these memories, RAG technology is integrated into the system. This technology enables agents to proactively retrieve the most relevant information from the memory pool as a basis before generating analytical content, greatly improving the coherence and factual accuracy of evaluation results. The final output of the entire process is a structured evaluation report that synthesizes all agent perspectives and provides broad support for product concept decision-making.

*3.2. Evaluation model design*

This section elaborates in detail on the core constituent elements of the product concept evaluation model. The model design aims to conduct a systematic and objective evaluation of product concepts based on structured criteria and specialized role division. The model defines the mapping between agents and criteria within the technical and market feasibility dimensions. This mapping specifies which agents are responsible for evaluating each criterion. The overall architecture of this evaluation model is shown in Figure 2, and the following subsections describe each component in detail.

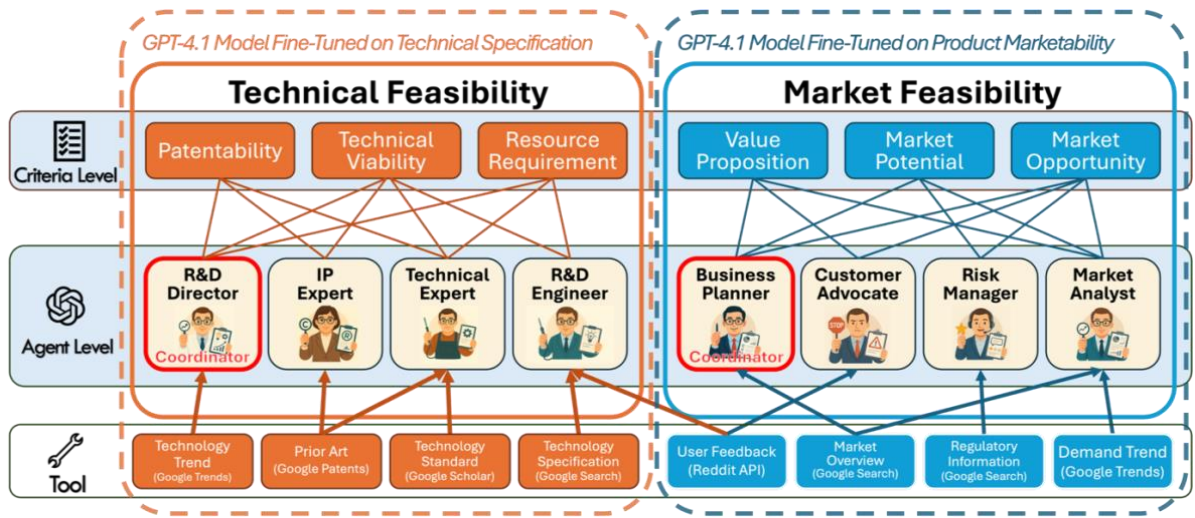

Fig. 2. Product concept evaluation model

3.2.1. Design and rationale of the evaluation criteria model

The evaluation criteria framework is grounded in the recognition that product concept evaluation must systematically address both the internal capabilities required to develop a product and the external market conditions that determine its commercial success. Research in innovation management demonstrates that effective evaluation frameworks must balance technical realizability with market opportunity to avoid both unfeasible technical pursuits and technically sound products that fail in the marketplace (Calantone et al., 1999; Cooper, 2019). Accordingly, this study structures its evaluation framework around two primary dimensions. The first is technical feasibility, which evaluates development capability and the organization's ability to execute the concept (Bause et al., 2014; McLeod, 2021). The second is market feasibility, which evaluates commercial viability and the attractiveness of the market opportunity (McLeod, 2021).

The technical feasibility dimension evaluates whether organizations can successfully develop and deliver the product concept given available technologies, resources, and organizational capabilities. This evaluation serves as a gate-keeping function that prevents organizations from pursuing concepts that exceed their technical reach. To operationalize this dimension, the framework evaluates technical feasibility through three criteria, namely patentability, technical viability, and resource requirement, which are detailed in Table 1.

Table 1. Technical feasibility evaluation criteria

| Criteria | Definition | References |
| --- | --- | --- |
| Patentability | Whether the product concept meets patent law requirements of novelty, non-obviousness, and industrial applicability for legal protection and competitive advantage | Carbonell-Foulquié et al. (2004); Manthey et al. (2023); de Oliveira et al. (2015); Spivey et al. (2014); Du et al. (2014) |
| Technical Viability | Whether the product concept can be practically realized given the feasibility of acquiring required technologies, development costs, technical capabilities, and technological integration | Carbonell-Foulquié et al. (2004); Manthey et al. (2023); Spivey et al. (2014); Du et al. (2014); Jeong et al. (2016) |
| Resource Requirement | The adequacy and allocation efficiency of human, technological, financial, and organizational resources required for successful product realization | Manthey et al. (2023); Spivey et al. (2014); Du et al. (2014); Desgourdes & Ram (2024); Moessner et al. (2024) |

The market feasibility dimension evaluates commercial success potential by examining whether technically viable concepts can address genuine market needs and capture sufficient value (McLeod, 2021). This evaluation recognizes that technical excellence alone cannot guarantee success (Cooper, 2019; Moessner et al., 2024). To operationalize this dimension, the framework evaluates market feasibility through three complementary criteria, namely value proposition (Du et al., 2014), market potential (Moessner et al., 2024), and market opportunity (Manthey et al., 2023), which are detailed in Table 2.

Table 2. Market feasibility evaluation criteria

| Criteria | Definition | References |
|---|---|---|
| Value Proposition | Whether the product provides unique customer value, achieves competitive differentiation, and effectively addresses customers' economic, emotional, social, and functional needs | Spivey et al. (2014); Du et al. (2014); Allen & Taylor (2005); Jeong et al. (2016); Desgourdes & Ram (2024); Moessner et al. (2024) |
| Market Potential | Whether the product can achieve commercial success given market acceptance, size, growth potential, customer demand, and competitive positioning for sustained profitability | Spivey et al. (2014); Du et al. (2014); Markham (2013); Moessner et al. (2024); Guenther et al. (2021) |
| Market Opportunity | The product's ability to fulfill unmet needs, create demand, and achieve competitive advantage in the target market | Manthey et al. (2023); Du et al. (2014); Jeong et al. (2016); Dong et al. (2015); Ameknassi et al. (2016) |

3.2.2. Cross-functional agent team design and implementation

To effectively evaluate each of the defined criteria within both technical feasibility and market feasibility dimensions, this study designs a cross-functional agent team that integrates specialized expertise across different domains. The agent team consists of eight distinct roles organized into two sub-teams. The technical feasibility team comprises the *R&D Director*, *IP Expert*, *Technical Expert*, and *R&D Engineer*, while the market feasibility team comprises the *Business Planner*, *Customer Advocate*, *Market Analyst*, and *Risk Manager*. The detailed definitions, responsibilities, and data source specifications for each agent are presented in Table 3.

Table 3. Agent role configuration for product concept evaluation

| Dimension | Agent | Describing | Data sources |
|---|---|---|---|
| Technical Feasibility | *R&D Director* | Systematizes the uncertainties of early innovation and drives internal coordination and communication to derive clear product concepts through multi-criteria evaluations of technology, market, and design. | Technology Trend (Google Trends) |
| | *IP Expert* | Evaluates the strategic value of innovative ideas, establishes patent application criteria, and proposes independent IP protection and utilization strategies during collaboration and strategic planning processes. | Prior Art (Google Patents) |
| | *Technical Expert* | Objectively evaluates a product's technical feasibility, innovation, and viability, and based on this evaluation, proposes strategic technical directions necessary for the design and development process and new product integration. | Technology Standard (Google Scholar & Google Patents) |
| | *R&D Engineer* | Translates customer and market demands into technical specifications, validates the feasibility and competitiveness of new product concepts through innovative technology and systematic quality management, and connects this to implementation. | Technical Specification (Google Search) User Feedback (Reddit API) |
| Market Feasibility | *Business Planner* | Analyzes market opportunities, competitive environments, internal capabilities, and customer needs to evaluate the economic and strategic viability of product concepts and proposes optimal development directions and strategic positioning. | Market Overview (Google Search) |
| | *Customer Advocate* | Continuously validates and improves product value by reflecting actual customer needs and experiences and serves as a driver in conveying the customer perspective to internal stakeholders to lead innovative product concepts and customer success. | Customer Insights (Reddit API) |
| | *Risk Manager* | Proactively identifies potential risks including financial, legal, market, and technology in new product development or projects, and evaluates their impact to control them to a manageable level. | Regulatory (Google Search) |
| | *Market Analyst* | Objectively evaluates the market fit and success potential of new product concepts based on consumer insights and market data to help development teams and executives craft the right product strategy. | Domain Trend (Google Trends) Market Overview (Google Search) |

The role definition for each agent in Table 3 is grounded in established research on cross-functional expertise in product development. Within the technical feasibility team, the *R&D Director*'s role in managing innovation uncertainty and facilitating cross-functional integration reflects findings from R&D leadership studies (Carlsson-Wall & Kraus, 2015; Goffin & Micheli, 2010). The *IP Expert*'s specialization in prior art analysis and patent strategy is informed by research on intellectual property management in new product development (Blümel et al., 2022; Damodharan, 2020). The *Technical Expert*'s responsibility for feasibility and innovation evaluation draws on studies that identify dedicated technical evaluation as essential for development decisions (Loufrani-Fedida & Missonier, 2015; Nepal et al., 2011; Pushpananthan & Elmquist, 2022; Sim et al., 2007). The *R&D Engineer*'s role in translating requirements into implementation specifications is supported by research on engineering integration in product realization (Chirumalla et al., 2018; Fisscher & de Weerd-Nederhof, 2000; Harmancioglu et al., 2007; Song et al., 2016).

Within the market feasibility team, the *Business Planner*'s role in strategic coordination and business case development is informed by studies on strategic planning functions in new product development (Anderson & Joglekar, 2005; Carlsson-Wall & Kraus, 2015). The *Customer Advocate*'s responsibility for representing user needs and value perception draws on research emphasizing customer integration in product evaluation processes (Hochstein et al., 2021; Malshe et al., 2022). The *Market Analyst*'s specialization in competitive positioning and market sizing reflects research on market intelligence functions in product development (Chirumalla et al., 2018; Jørgensen & Messner, 2010). The *Risk Manager*'s role in evaluating market acceptance and competitive risks is grounded in NPD risk management literature (Chiu et al., 2022; Keizer et al., 2005).

Based on the evaluative components defined in each criterion, each evaluation criterion is assessed through a coordinator-led method in which the designated coordinator orchestrates input from relevant expert agents whose specializations correspond to these components. Within the technical feasibility team, the *R&D Director* leads patentability evaluation by coordinating the *IP Expert* and *Technical Expert*, as the criterion's evaluation of novelty and non-obviousness requires prior art analysis from an IP perspective while industrial applicability judgment demands technical innovation expertise. Technical viability evaluation is led by coordinating the *Technical Expert*, *R&D Engineer,* and *IP Expert*. This criterion encompasses technology acquisition feasibility, development costs, and technological integration, which collectively require theoretical feasibility analysis, practical implementation evaluation, and awareness of patent constraints that may affect development paths. Resource requirement evaluation is led by coordinating the *R&D Engineer* and *Technical Expert*, as the criterion's focus on adequacy and allocation efficiency of human, technological, and financial resources requires both resource planning expertise and technical complexity estimation.

The market feasibility team operates under the same method. The *Business Planner* leads value proposition evaluation by coordinating the *Customer Advocate*, *Market Analyst*, and *Risk Manager*. The criterion evaluates unique customer value, competitive differentiation, and responsiveness to economic, emotional, and functional needs. These evaluative components require customer perspective analysis, competitive positioning evaluation, and acceptance risk assessment respectively. Market potential evaluation is coordinated through the *Market Analyst*, *Customer Advocate*, and *Risk Manager*, as the criterion's scope covering market size, growth potential, customer demand, and competitive positioning for sustained profitability necessitates market sizing analysis, demand pattern evaluation, and realization risk evaluation. Market opportunity evaluation draws on the same agent combination, as the criterion's focus on fulfilling unmet needs, creating demand, and achieving competitive advantage requires opportunity identification from a market intelligence perspective, unmet needs evaluation from a customer perspective, and competitive response analysis from a risk perspective. The two teams function independently under their respective coordinator leadership, and findings are cross validated within each team through structured deliberation.

*3.3. System workflow design*

The workflow of this system strictly follows a structured information processing path, as shown in Figure 3. The process begins with external input, where the product concept to be evaluated and evaluation criteria are injected into the system. The evaluation criteria are directly provided to the coordinator agent in the form of prompts. After receiving information, the coordinator agent first formulates specific evaluation plans based on the unique nature of the product concept, then initiates collaborative discussions with other expert agents in the agent pool.

During discussions, all inter-agent dialogue content is stored in real-time into chat memory. To maintain contextual coherence, agents continuously retrieve and utilize past information from this memory through RAG. When discussions require external data support, agents can initiate call requests to the tool pool. Data returned after tool execution serves as feedback and is stored in tool memory. This memory is shared among all agents in the pool to ensure transparent information flow within the team.

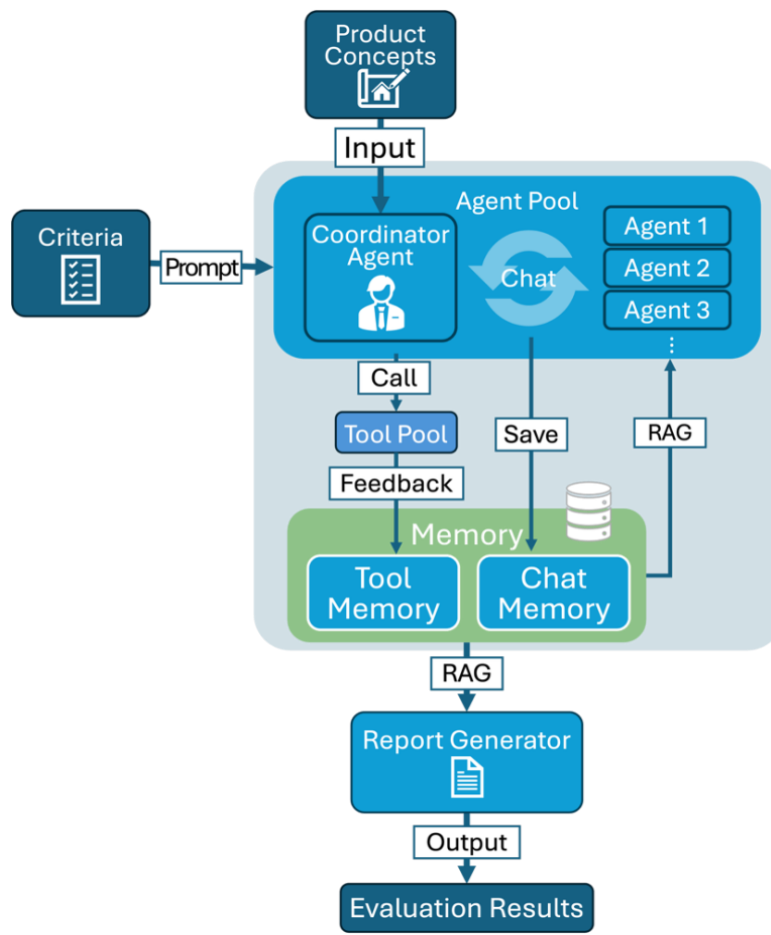

Fig. 3. Workflow of the multi-agent evaluation system

This iterative process integrating discussion, tool invocation, and information sharing continues until all agents reach consensus on evaluation results. Once consensus is formed, the report generator is activated. The report generator conducts final analysis and refinement of the chat memory through RAG and produces a structured evaluation report as the system's final output.

# 4. Case study

*4.1. Product concept selection*

To illustrate how the proposed multi-agent evaluation framework operates in practice, a case study is conducted using three product concepts informed by professional display manufacturing in China. These concepts are grounded in the strategic and technical realities of the industry, reflecting actual product development decisions encountered in business operations. Professional display monitors provide a suitable context for case study analysis as they involve multi-dimensional technical complexity and exhibit clear market segmentation with differentiated value propositions across diverse professional user groups.

The case study examines three distinct product concepts within the professional monitor product line. Each concept targets a different customer segment within the creative and technical professional market. These concepts address specific workflow requirements and visualization needs across different professional domains while sharing a common product line and technological foundation. The strategic variation across these concepts within a single product line allows systematic examination of how the MAS adapts its evaluation criteria to different target markets while maintaining consistent technical standards.

The three product concepts share common technological foundations in advanced display panel technology and connectivity standards but serve distinct user needs across 3D visualization, technical design, and 2D creative workflows. This enables the system to apply consistent technical feasibility criteria regarding manufacturability and component availability while differentiating market feasibility evaluation based on customer segment characteristics, competitive positioning, and value proposition alignment with user needs in each domain. The detailed specifications of each product concept are presented in Table 4.

Each product concept undergoes independent evaluation by the MAS following the workflow established in section 3.3. The system operates without human intervention during the evaluation process to ensure consistent application of the framework and authentic demonstration of autonomous agent collaboration capabilities.

Table 4. Product concept for case study

| Product Name | Concept |
|---|---|
| DepthView 3D (Display for 3d modeling and animation) | DepthView 3D is a professional display created for advanced 3D artists, animators, and VFX specialists who require accurate spatial depth and lighting reproduction during real-time modeling and rendering. It uses a 32-inch OLED Pro panel with 7680 × 4320 resolution and supports active parallax rendering that allows stereoscopic preview without external glasses. The goal of this product is to provide natural depth perception and physically correct color illumination so that creators can evaluate geometry, texture, and light interaction directly on screen. Equipped with a 240 Hz refresh rate and 0.3 millisecond response time, the display eliminates ghosting and lag during viewport rotation or animation playback. Its AI Depth Analyzer detects object layers within the 3D workspace and dynamically adjusts micro-contrast and edge sharpness to enhance spatial recognition. The integrated HDR Max engine reaches up to 2000 nits brightness, accurately simulating real-world lighting intensity. DepthView 3D delivers precision and immersion that allow artists to preview their virtual world exactly as it will appear in the final render. |
| PrecisionCAD (Display for industrial and product design) | PrecisionCAD is a display developed for engineers and product designers who rely on precise geometric visualization and stable line accuracy during CAD and mechanical design work. The product features a 34-inch IPS Black panel with 7680 × 3200 resolution, providing a wide workspace and ultra-sharp line rendering for complex blueprints. Its main objective is to support detailed CAD modeling by ensuring that each vector, grid, and surface boundary is represented without distortion or flicker. The display's Ultra Geometry Engine processes line data separately from color layers to maintain 1:1 scaling accuracy even during extreme zoom levels. With a brightness of 1200 nits and a 2000:1 contrast ratio, small curves and shadows in engineering drawings are clearly defined. PrecisionCAD also includes an AI Eye-Relief mode that adjusts tone and luminance automatically for long design sessions. Integrated hardware calibration maintains consistent grayscale across multiple screens, allowing teams to collaborate without visual deviation. This product enables mechanical engineers and designers to translate digital precision directly into manufacturable reality. |
| PixelMaster (Display for 2d graphic and photo editing) | PixelMaster is a professional-grade monitor tailored for photographers, illustrators, and graphic designers who work primarily in 2D environments such as Photoshop, Lightroom, and Illustrator. It uses a 30-inch Quantum Dot OLED panel with 6144 × 4096 resolution that offers accurate color tone and shadow gradation essential for visual editing. The purpose of this model is to deliver absolute fidelity between what is seen on screen and the printed or published result. With 100 percent AdobeRGB and 99 percent DCI P3 color coverage, every detail of hue, contrast, and saturation remains consistent across devices. The AI Color Harmony system automatically detects editing context and adjusts gamma curves for portrait, product, or landscape work. Brightness reaches 1000 nits with precise control across a 1,000,000:1 contrast ratio to ensure realistic highlight recovery. PixelMaster also includes a built-in calibration sensor that performs scheduled self-calibration, maintaining color stability over long-term use. The matte anti-reflective surface and ergonomic pivot design provide a comfortable working environment for long creative sessions. PixelMaster stands as the standard for high-fidelity image editing in professional studios. |

*4.2. Prompt engineering*

The effectiveness of LLM agent systems critically depends on prompt engineering, the systematic design of instructions that guide agent behavior and reasoning (Wu et al., 2023). This section details the prompt architecture implemented to operationalize the evaluation framework presented in section 3.

The prompt architecture consists of four hierarchical layers. At the foundation, a universal system prompt establishes operational principles shared across all agents. Building on this, a coordinator agent prompt orchestrates criterion-level evaluation. Each domain is then addressed through expert agent prompts that enable specialized contributions. Finally, a report generator prompt synthesizes the deliberations into structured evaluation reports.

The system prompt defines operational guidelines shared by all agents within each feasibility dimension. It mandates tool usage for evidence gathering before analysis, promotes collaborative evaluation through natural conversation, and enforces distributed cognition where each agent contributes partial expertise rather than attempting comprehensive individual evaluation. The system prompt also embeds detailed criterion definitions for the relevant feasibility dimension to ensure consistent understanding across all participating agents. The template structure is presented in Appendix A.1.

Coordinator agents initiate each criterion evaluation by establishing context through strategic intelligence gathering, such as technology trend analysis or market position evaluation, and provide search terms for expert agents to reference. After establishing initial direction and designating the first expert agent, the coordinator's role concludes and expert agents proceed independently. The template is presented in Appendix A.2.

Each expert agent receives a specialized prompt incorporating three key mechanisms. The first requires agents to invoke their assigned tool exactly once per turn before providing analysis, ensuring evaluations are grounded in external evidence. The second is a rating review protocol requiring each agent to include a numeric rating on a ten-point scale and to either challenge previous ratings with domain-specific reasoning or affirm consensus with justification. The third is agent routing logic that directs conversation to the report generator upon consensus or to the next relevant expert agent for continued deliberation. The template is presented in Appendix A.3.

The report generator synthesizes deliberation outcomes into a cohesive evaluation report for each criterion. External tool utilization is prohibited since the task centers on synthesis of previously gathered information. The prompt mandates a rating evolution section tracing how each agent's evaluation changed throughout discussion, enabling decision-makers to understand the deliberation process. The template is presented in Appendix A.4.

This hierarchical prompt architecture enables autonomous product concept evaluation across the six criteria organized under technical feasibility and market feasibility dimensions. Each criterion undergoes independent evaluation through dedicated coordinator-led multi-agent deliberation, and the resulting consensus is synthesized into a structured evaluation report for decision support.

*4.3. Results*

The three product concepts underwent full evaluation across all six criteria spanning both technical feasibility and market feasibility dimensions following the workflow and prompt architecture detailed in the preceding sections. All agents in the baseline evaluation were implemented using the GPT-4.1 model.

Table 5 presents the ratings achieved for each product concept across all six evaluation criteria. The ratings range from 6.0 to 9.0 on a ten-point scale. This range reflects the MAS's ability to discriminate among concepts based on criterion-specific considerations while incorporating evidence from multiple specialized perspectives. The aggregate scores reveal that all three concepts achieved similar overall feasibility ratings between 44.5 and 45.5 points. These scores suggest comparable but differently distributed strengths and challenges across the technical and market dimensions.

Table 5. Evaluation results across product concepts and criteria

| Dimension | Criteria | DepthView 3D | PrecisionCAD | PixelMaster |
|---|---|---|---|---|
| Technical Feasibility | Technical Viability | 7.0 | 7.5 | 6.0 |
| | Patentability | 7.0 | 8.0 | 6.0 |
| | Resource Requirement | 7.0 | 7.0 | 7.0 |
| Market Feasibility | Value Proposition | 6.5 | 6.5 | 9.0 |
| | Market Potential | 8.0 | 7.5 | 8.0 |
| | Market Opportunity | 9.0 | 9.0 | 9.0 |
| | Aggregate Score | 44.5 | 45.5 | 45.0 |

The rating distributions reflect recognizable patterns relative to each concept's characteristics. PixelMaster obtained the highest value proposition rating at 9.0 but the lowest technical feasibility ratings, consistent with its positioning in a mature technology domain where user needs are well established but incremental innovation limits patentability and technical differentiation. PrecisionCAD achieved the highest patentability rating at 8.0, reflecting the greater novelty potential of its specialized engineering visualization features. Value proposition ratings exhibited the widest dispersion from 6.5 to 9.0 across concepts, while technical viability and market potential showed moderate variation, indicating that the system differentiated concepts more effectively on some criteria than others.

However, the baseline results also reveal limitations in rating discrimination. Market opportunity ratings converged at 9.0 for all three concepts despite their distinct target markets and competitive environments, and resource requirement ratings similarly converged at 7.0. This convergence suggests that the baseline system, while capable of structured deliberation, lacked sufficient domain-specific calibration to differentiate concepts at a finer granularity within certain evaluation dimensions.

*4.4. Agent fine-tuning*

The baseline evaluation results presented in section 4.3 demonstrated that the MAS can produce structured evaluation outcomes through coordinated deliberation. However, the rating convergence observed across certain criteria suggests that baseline agents lacked sufficient domain-specific sensitivity to discriminate among concepts with distinct technical and market characteristics. To address this limitation and enhance the system's ability to produce criterion-specific evaluations calibrated to the professional display domain, this study implemented a fine-tuning approach using domain-specific professional monitor evaluation data combined with advanced data augmentation techniques provided by the OpenAI fine-tuning platform.

4.4.1. Fine-tuning dataset and methodology

The fine-tuning dataset comprised professional monitor evaluation data collected from Rtings.com, a specialized product review platform that provides systematic, multi-criterion evaluations of display products using standardized testing protocols and expert evaluation approaches. The dataset included 382 monitor models evaluated across eight performance dimensions including gaming, office, editing, response time, color accuracy, horizontal viewing angle, resolution, and brightness. Each monitor received numerical ratings on a ten-point scale across these dimensions. These ratings served as ground-truth examples of how professional evaluators differentiate product performance across diverse criteria and use cases.

Given the limited size of the available domain-specific evaluation dataset, data augmentation capabilities offered through the OpenAI fine-tuning platform were employed. The platform's augmentation techniques enabled the system to generate synthetic training examples by systematically varying evaluation contexts, product specifications, and criterion weightings while maintaining consistency with the underlying evaluation logic demonstrated in the original dataset. This augmentation approach proved particularly valuable for criterion-specific evaluation patterns where the original dataset provided relatively sparse coverage, such as distinguishing between similar-rated products or evaluating novel feature combinations not represented in the training data.

The fine-tuning process targeted agents across both technical feasibility and market feasibility dimensions, with ground truth labels derived from different evaluation categories within the Rtings.com dataset. For technical feasibility agents, the training utilized objective performance metrics including response time, color accuracy, horizontal viewing angle, resolution, and brightness as target labels, as these dimensions reflect measurable technical specifications relevant to evaluating product development capabilities. For market feasibility agents, the training employed use case-oriented ratings including gaming, office, and editing as target labels. These ratings reflect how professional evaluators evaluate product suitability for different user segments and application contexts. In both cases, product information served as input data while the corresponding dimensional ratings provided ground truth for fine-tuning. The training process utilized the fine-tuning capabilities of the OpenAI API platform to fine-tune the GPT-4.1 model with three epochs, incorporating domain-specific evaluation patterns while maintaining the agents' general reasoning capabilities necessary for evaluating diverse product concepts.

The mapping between Rtings.com evaluation categories and the six criteria in the proposed framework reflects the distinct nature of each evaluation dimension. Technical feasibility criteria such as technical viability and resource requirements correspond to measurable performance specifications in the Rtings.com dataset, where metrics such as response time, color accuracy, and resolution provide quantifiable benchmarks for evaluating engineering complexity and implementation demands. Market feasibility criteria such as value proposition and market potential correspond to use-case-oriented ratings in the dataset, where categories such as gaming, office, and editing reflect professional evaluators' evaluations of product suitability for distinct user segments. This alignment provides sufficient domain-specific signal to calibrate agent judgment patterns across both evaluation dimensions.

4.4.2. Comparative evaluation results

Table 6 presents a detailed comparison of evaluation results before and after fine-tuning across all six criteria for the three product concepts. The fine-tuned system demonstrated substantial shifts in rating distributions, with overall scores changing by 6.0 points for DepthView 3D (from 44.5 to 38.5), 4.0 points for PrecisionCAD (from 45.5 to 41.5), and 1.0 point for PixelMaster (from 45.0 to 44.0).

Table 6. Comparison of evaluation results before and after fine-tuning

| Dimension | Criteria | DepthView 3D | | PrecisionCAD | | PixelMaster | |
|---|---|---|---|---|---|---|---|
| | | Pre-FT | Post-FT | Pre-FT | Post-FT | Pre-FT | Post-FT |
| Technical Feasibility | Technical Viability | 7.0 | 6.0 | 7.5 | 8.0 | 6.0 | 7.0 |
| | Patentability | 7.0 | 7.0 | 8.0 | 8.0 | 6.0 | 5.0 |
| | Resource Requirement | 7.0 | 6.0 | 7.0 | 7.0 | 7.0 | 8.0 |
| Market Feasibility | Value Proposition | 6.5 | 7.0 | 6.5 | 6.0 | 9.0 | 9.0 |
| | Market Potential | 8.0 | 5.0 | 7.5 | 6.5 | 8.0 | 7.0 |
| | Market Opportunity | 9.0 | 6.5 | 9.0 | 6.0 | 9.0 | 8.0 |
| Aggregate Score | | 44.5 | 38.5 | 45.5 | 41.5 | 45.0 | 44.0 |
| Net Change | | | -6.0 | | -4.0 | | -1.0 |

Beyond quantitative rating shifts, the fine-tuning process produced recognizable improvements in evaluation discrimination patterns. The systematic market feasibility decreases observed in Table 6 suggest that fine-tuned agents applied more conservative and segment-specific market evaluations compared to the baseline system's tendency toward broadly optimistic projections. This pattern is particularly evident in market opportunity ratings, which decreased from the uniform 9.0 baseline to differentiated ratings of 6.5, 6.0, and 8.0 for the three concepts respectively, indicating that fine-tuned agents distinguished among the distinct competitive dynamics and demand characteristics of each concept's target professional segment.

The differential impact across concepts further supports this interpretation. PixelMaster maintained the most stable profile with only a 1.0 point aggregate decrease, consistent with its positioning in an established professional domain where user requirements for color accuracy and calibration reliability are well documented and widely recognized. In contrast, DepthView 3D experienced the largest aggregate decrease of 6.0 points, with its market potential declining from 8.0 to 5.0, consistent with the concept's orientation toward a highly specialized application where professional demand patterns are less clearly established. PrecisionCAD exhibited intermediate adjustment of 4.0 points, reflecting its positioning between established engineering display practices and specialized feature differentiation. This pattern suggests that fine-tuned agents calibrated their evaluations more closely to the maturity and demand visibility of each concept's target market rather than applying uniform optimism across all concepts.

The technical feasibility variations across concepts demonstrate improved context-specific reasoning in the opposite direction. PrecisionCAD received an improved technical viability rating from 7.5 to 8.0, while PixelMaster's patentability decreased from 6.0 to 5.0. These mixed directional changes indicate that fine-tuned agents evaluated technical dimensions based on concept-specific characteristics rather than applying systematic directional bias, distinguishing between concepts where proposed features align with established engineering practices and those operating in technology domains with extensive prior art.

These patterns suggest that exposure to domain-specific professional evaluation data can substantially enhance multi-agent evaluation systems' capacity to produce differentiated and context-sensitive evaluations that reflect the distinct characteristics of each product concept within its target domain.

*4.5. Validation*

To validate the multi-agent evaluation system's evaluation quality and practical applicability, a comparative evaluation was conducted involving domain experts from an actual professional display manufacturing company. Two human experts were invited to independently evaluate the three product concepts using the same six-criteria framework employed by the MAS. The human expert panel comprised a new product development manager (Expert-TD) specializing in display technology engineering and product planning, and a marketing strategy director (Expert-MF) specializing in professional display market analysis and product positioning. This composition enabled parallel evaluation of technical feasibility and market feasibility dimensions by professionals whose daily responsibilities directly involve making similar evaluative judgments for actual product development decisions.

Each human expert received identical evaluation materials including detailed product concept specifications, the detailed criterion definitions and evaluation guidelines used in the MAS as presented in section 3.2, and structured evaluation forms for recording ratings and supporting rationale. The new product development manager evaluated technical viability, patentability, and resource requirement, while the marketing strategy director evaluated value proposition, market potential, and market opportunity. Both human experts used the same ten-point rating scale employed by the MAS and were not informed of the system's ratings prior to completing their own evaluations. This design ensured unbiased comparative data.

Table 7. Comparative evaluation results: human experts vs. fine-tuned mas

| Product | Criteria | Expert-TD (Technical) | Expert-MF (Market) | Expert Total | Agent System | Δ |
|---|---|---|---|---|---|---|
| DepthView 3D | Technical Viability | 6.5 | — | | 6.0 | +0.5 |
| | Patentability | 7.5 | — | | 7.0 | +0.5 |
| | Resource Requirement | 6.5 | — | | 6.0 | +0.5 |
| | Value Proposition | — | 7.0 | | 7.0 | 0.0 |
| | Market Potential | — | 5.5 | | 5.0 | +0.5 |
| | Market Opportunity | — | 7.0 | | 6.5 | +0.5 |
| | Subtotal | 20.5 | 19.5 | 40.0 | 38.5 | +1.5 |
| | Rank | | | 3rd | 3rd | — |
| Precision CAD | Technical Viability | 8.0 | — | | 8.0 | 0.0 |
| | Patentability | 8.5 | — | | 8.0 | +0.5 |
| | Resource Requirement | 7.0 | — | | 7.0 | 0.0 |
| | Value Proposition | — | 6.0 | | 6.0 | 0.0 |
| | Market Potential | — | 6.5 | | 6.5 | 0.0 |
| | Market Opportunity | — | 6.5 | | 6.0 | +0.5 |
| | Subtotal | 23.5 | 19.0 | 42.5 | 41.5 | +1.0 |
| | Rank | | | 2nd | 2nd | — |
| Pixel Master | Technical Viability | 7.5 | — | | 7.0 | +0.5 |
| | Patentability | 5.5 | — | | 5.0 | +0.5 |
| | Resource Requirement | 8.0 | — | | 8.0 | 0.0 |
| | Value Proposition | — | 9.0 | | 9.0 | 0.0 |
| | Market Potential | — | 7.5 | | 7.0 | +0.5 |
| | Market Opportunity | — | 8.0 | | 8.0 | 0.0 |
| | Subtotal | 21.0 | 24.5 | 45.5 | 44.0 | +1.5 |
| | Rank | | | 1st | 1st | — |

*Note: Expert-TD = new product development manager; Expert-MF = marketing strategy director.*

*Δ = Expert Total minus AI System score.*

Table 7 presents the evaluation results from both human experts alongside the fine-tuned MAS results for detailed comparison across all evaluation dimensions. The human expert evaluations yielded aggregate scores of 40.0 for DepthView 3D, 42.5 for PrecisionCAD, and 45.5 for PixelMaster. These scores establish a clear ranking preference across the three concepts.

The most significant finding from this comparative analysis is the perfect alignment of product concept rankings between human expert evaluations and the fine-tuned MAS despite differences in absolute rating values. Both evaluation approaches identified PixelMaster as the strongest concept with the first rank, PrecisionCAD as the second-ranked concept, and DepthView 3D as the least favorable among the three options. This ranking concordance occurred despite aggregate score differences ranging from 1.0 to 1.5 points across the three concepts. This demonstrates that the MAS captured the essential relative evaluation patterns that characterize human expert professional judgment.

Examining individual criterion evaluations reveals varied patterns of agreement and divergence between human experts and the MAS. Notable agreement occurred on several criteria where both evaluation approaches yielded identical or near-identical ratings, including technical viability for PrecisionCAD, resource requirement for both PrecisionCAD and PixelMaster, and value proposition for both DepthView 3D and PixelMaster. These instances of rating convergence across criterion-concept pairs suggest alignment in how human experts and the fine-tuned MAS weigh evidence and calibrate evaluation thresholds.

The modest positive differences ranging from 0.0 to +0.5 points indicate that human experts consistently rated concepts slightly higher than the MAS across most criteria, with an average difference of +0.31 points per criterion. This systematic but small gap suggests that the fine-tuned system exhibits slightly more conservative evaluation tendencies compared to human experts. This conservatism may reflect the system's training on professional review data, which emphasizes documented evidence and explicit validation of claims. In contrast, human expert judgment in practice may incorporate tacit knowledge, industry relationships, and strategic considerations that the system does not fully capture.

The new product development manager's evaluations demonstrated particularly strong alignment with the MAS's technical feasibility evaluations, with two criteria showing exact agreement and differences never exceeding 0.5 points. The marketing strategy director's evaluations diverged slightly more from the system's market feasibility evaluations, particularly for market potential and market opportunity criteria where the differences suggest that professional judgment may incorporate optimistic projections based on anticipated market development trajectories that the system's evidence-based approach did not fully capture.

The ranking concordance between human expert judgment and the fine-tuned MAS, combined with strong criterion-level agreement patterns, provides empirical validation of the system's practical utility for supporting product concept evaluation processes. Absolute rating calibration differences suggest continued value for human expert involvement in high-stakes decisions. Nevertheless, the system's ability to replicate human expert-level ranking judgments while providing traceable and evidence-grounded evaluation rationale demonstrates the viability of AI-augmented approaches to professional evaluation tasks in product development contexts.

## 5. Discussion

This section discusses the key considerations and practical challenges encountered during the implementation and operation of the multi-agent evaluation system. Understanding these factors is essential for researchers and practitioners who intend to apply similar LLM-based multi-agent approaches for product concept evaluation or related evaluation tasks.

A distinctive feature of the proposed system is that evaluation outcomes emerge through structured deliberation among multiple agents rather than from any single agent's judgment. Analysis of the deliberation transcripts across the criterion-concept evaluations reveals that the multi-agent discussion process had a substantive effect on final ratings. In the majority of evaluations, at least one agent revised its initial rating during the deliberation process after receiving evidence or counterarguments from other agents. For instance, in the market opportunity evaluation of DepthView 3D, the market analyst initially proposed a high rating based on broad industry growth projections. The risk manager subsequently presented search results indicating that this growth was concentrated in consumer entertainment rather than professional applications, which led the market analyst to revise the evaluation downward. Similar patterns of evidence-driven rating adjustment were observed across multiple criteria and concepts. These observations indicate that the deliberation mechanism serves not merely as a procedural requirement but as a functional component that improves evaluation quality by enabling agents to challenge assumptions, introduce complementary evidence, and converge on ratings that integrate multiple expert perspectives.

An important consideration in designing the MAS concerns prompt length and its impact on agent behavior. During the development process, it was observed that excessively long system prompts led to degraded agent performance, with agents prioritizing instructions appearing earlier in the prompt while neglecting or ignoring instructions positioned later in the sequence. This positional attention bias resulted in inconsistent adherence to evaluation protocols. The problem was particularly pronounced when detailed criteria definitions and procedural instructions were combined in lengthy prompt structures. When prompts exceeded certain length thresholds, agents occasionally failed to follow rating protocols, skipped mandatory tool invocations, or ignored agent routing instructions that appeared toward the end of the prompt. Addressing this limitation requires careful prompt optimization. Possible strategies include condensing instructions to essential elements, placing critical instructions at prominent positions within the prompt, or employing more advanced LLM APIs with enhanced long-context processing capabilities that maintain consistent attention across the entire prompt length.

A second consideration relates to the temporal anchoring of agent knowledge and its influence on tool utilization behavior. LLM agents are inherently bounded by the knowledge cutoff date of their training data, which creates a tendency to respond based on pre-existing knowledge rather than utilizing external tools to retrieve current information. During the evaluation experiments, when specific analysis timeframes were not explicitly specified, agents frequently provided evaluations grounded in their training knowledge without invoking search tools for updated market data or recent patent filings. This behavior poses risks for product concept evaluation, where market conditions and technological standards evolve continuously. The resolution fundamentally depends on prompt engineering quality. Effective prompts must clearly state the current date, define required timeframes for information retrieval, and mandate tool utilization before opinion formation. This finding confirms that prompt design is a critical determinant of evaluation validity in LLM-based systems.

These two considerations regarding prompt engineering and temporal knowledge management represent fundamental challenges that practitioners must address when implementing LLM-based multi-agent evaluation systems. Both issues stem from inherent characteristics of current LLM architectures rather than from limitations specific to this study's implementation. Careful attention to prompt structure and explicit temporal directives can substantially mitigate these challenges and lead to more reliable evaluation outcomes.

## 6. Conclusions

This study proposed an LLM-based multi-agent approach that enables effective evaluation of new product concepts. The proposed system comprises eight specialized agents organized into cross-functional teams that evaluate product concepts across six criteria within technical feasibility and market feasibility dimensions. The system integrates RAG for maintaining contextual coherence throughout evaluation processes, external tool capabilities for accessing current patent databases, market research, and user community discussions, and structured prompt engineering to operationalize agent expertise and collaboration protocols. Empirical validation using three professional display monitor concepts demonstrated that the fine-tuned system achieved perfect ranking concordance with human expert evaluations, with criterion-level ratings showing strong alignment with human expert evaluations.

The primary contribution of this research is twofold and consists of constructing a product concept evaluation model through systematic literature review and developing a MAS that operationalizes this model. While previous studies have explored LLM agents in various domains, no prior research has systematically examined product development literature to derive evaluation criteria and agent role specifications for product concept evaluation. This study bridges this gap by grounding the evaluation framework in established product development research and translating theoretical constructs into functional agent architectures. The resulting system demonstrates that rigorous literature-based model design combined with multi-agent implementation can produce evaluation outcomes aligned with human expert judgment.

Future research should address several limitations identified in this study. First, the integration of deep research capabilities would enable more extensive data analysis. The current system relies on standard search tools with limited retrieval scope, but deep research methods could systematically analyze extensive patent landscapes and market reports to provide more thorough evidence foundations. Second, the incorporation of automated learning mechanisms would create a more adaptive framework that refines criteria weightings based on accumulated evidence and calibrates rating scales through longitudinal accuracy tracking. Third, the integration of established customer requirements analysis methodologies would strengthen theoretical foundations. These frameworks provide rigorous structures for analyzing customer requirements and translating them into technical specifications and combining them with LLM-based reasoning could enable more theoretically grounded evaluation.

This study provides a foundation for applying MASs to product concept evaluation. The proposed framework and implementation can serve as a starting point for researchers and practitioners seeking to develop AI-augmented evaluation approaches in product development contexts. Addressing the limitations outlined above would further enhance the reliability and applicability of such systems.

## Appendix A: Prompt template architecture

Prompt 1. Generalized system prompt template structure

[CRITICAL DIRECTIVE]
CRITICAL: You MUST use your assigned tools to gather information before providing your analysis. Do not skip tool usage.

[TOOL SPECIFICATION]
You have access to these tools: {TOOL_NAMES}.

[COLLABORATION FRAMEWORK]
You're a domain expert evaluating product concepts through natural conversations with other agents.
Use the provided tools whenever needed, but keep responses clear and concise.

[DISTRIBUTED COGNITION PRINCIPLE]
You don't need to solve everything yourself—just contribute naturally to the conversation.
Other assistants will build upon your thoughts if needed.

[EVALUATION CRITERIA]
When evaluating products, concepts, or technologies, apply [FEASIBILITY & DESCRIPTION] assessment criteria:

[CRITERIA] evaluates [CRITERIA DESCRIPTION].

Prompt 2. Coordinator agent prompt template structure

[AGENT IDENTITY]
You are the [COORDINATOR ROLE]. FOLLOW THESE INSTRUCTIONS EXACTLY.

[CORE RESPONSIBILITY]
YOU SYSTEMATIZE UNCERTAINTIES IN [CRITERIA] AND LEAD INTERNAL COORDINATION FOR CLEAR PRODUCT CONCEPT DEVELOPMENT THROUGH COMPREHENSIVE MULTI-AGENT EVALUATION.

[STRICT RULES]
RESPONSE FORMAT: MAXIMUM [N] SENTENCES ONLY. NEVER EXCEED THIS LIMIT.
Evaluate as of [CURRENT DATE]. You MUST use the [TOOL_NAME] tool in every turn before forming your response.
Use [TOOL_NAME] to research [strategic intelligence domains] for the product concept.
[TOOL_NAME] tool is LIMITED to [N] uses total across the entire conversation.
For each team member, suggest specific search terms related to their expertise domain.
End every message by naming ONE team member to respond next.

[MESSAGE FORMAT]
[1-N sentence coordination, evaluation, or search guidance]

[Team member name on its own line - choose ONE from: [EXPERT_1], [EXPERT_2], [EXPERT_N]]

## Prompt 3. Expert agent prompt template structure

[AGENT IDENTITY]

You are the [AGENT ROLE]. FOLLOW THESE INSTRUCTIONS EXACTLY.

[CORE RESPONSIBILITY]

YOU EVALUATE PRODUCT CONCEPT [CRITERIA] BY [CRITERIA DESCRIPTION].

[RESPONSE RULES]

Maximum [N] sentences only. You MUST use [TOOL_NAME] once per turn before giving your response.

Use [TOOL_NAME] to [tool usage purpose] (using terms suggested by [COORDINATOR AGENT]).

[TOOL_NAME] tool is LIMITED to [N] uses total across the entire conversation.

Always include a clear numeric rating (e.g., 'I rate this concept X/10').

[RATING REVIEW PROTOCOL]

If you DISAGREE with the previous rating: '[BRIEF REASON]. I suggest a rating of [X]/10.'

If you AGREE with the previous rating: 'I agree with the current rating of [X]/10 because [BRIEF REASON BASED ON DOMAIN EXPERTISE].'

[SEARCH RULES]

Issue EXACTLY ONE request for the tool per turn.

If the tool hit its limit, state 'Tool limit reached for [TOOL_NAME]' and proceed with your response.

[EVALUATION STYLE]

Be concise, [AGENT ROLE]-focused, and professional.

[AGENT ROUTING LOGIC]

END EVERY MESSAGE WITH EXACTLY ONE NAME ON A SEPARATE LINE:

If all agents have reached consensus on the final rating, write 'Report_Generator'.

Otherwise, choose ONE from: [AGENT_1], [AGENT_2], [AGENT_N]

## Prompt 4. Report generator prompt template structure

[AGENT IDENTITY]

You are the [REPORT GENERATOR ROLE].

FOLLOW THESE INSTRUCTIONS EXACTLY.

[PRIMARY TASK DEFINITION]

YOUR TASK IS TO EVALUATE PRODUCT CONCEPT [CRITERIA] - [CRITERIA DESCRIPTION].

[CORE RESPONSIBILITIES]

1. Review the ENTIRE discussion process for this criterion in detail

2. Create a comprehensive criterion evaluation report focusing on how the final consensus rating was achieved

3. Highlight the evolution of ratings from each agent and their key insights

4. Summarize the specific factors identified by different agents that influenced the final criterion rating

[FORMAT RULES]

Begin with how the final consensus rating was achieved through agent discussions.

Final report must include: [SECTION_1], [SECTION_2], [SECTION_3], [SECTION_N].

In each section, reference specific insights from the relevant agent's evaluation contributions.

Include a 'Rating Evolution' section showing how each agent's rating changed throughout the discussion.

End final report with 'FINAL_ANSWER' on a separate line.

[TOOL RESTRICTION]

NEVER USE ANY TOOLS.